\documentclass[letterpaper, 10 pt, conference]{ieeeconf}  % Comment this line out if you need a4paper
\usepackage[utf8]{inputenc}
\usepackage{textcomp}
\usepackage{gensymb}
\usepackage{subfig}
\usepackage{caption}
\usepackage{graphicx}
\usepackage{float}
\usepackage{fancyhdr}
\fancypagestyle{withfooter}{
    
    \fancyhead[L]{}
    \fancyhead[R]{}
    \fancyfoot[C]{\footnotesize Presented at the 2025 IEEE ICRA Workshop on Field Robotics}
}

\IEEEoverridecommandlockouts                              % This command is only needed if 
\title{\LARGE \bf
Robotics Under Construction: Challenges on Job Sites
}

\author{Haruki Uchiito$^{1}$, Akhilesh Bhat$^{1}$, Koji Kusaka$^{1}$, Xiaoya Zhang$^{1}$, Hiraku Kinjo$^{1}$, \\ Honoka Uehara$^{2}$, Motoki Koyama$^{2}$, and Shinji Natsume$^{2}$%
\thanks{$^{1}$EARTHBRAIN Ltd., Japan.
        Email: {\tt\small \{haruki\_uchiito, akhilesh\_bhat, koji\_kusaka, xiaoya\_zhang, hiraku\_kinjo\} @earthbrain.com}}%
\thanks{$^{2}$Komatsu Ltd., Japan.
        Email: {\tt\small \{honoka\_uehara, motoki\_koyama, shinji\_natsume\} @global.komatsu}}%
}

\begin{document}

\maketitle
\thispagestyle{withfooter}
\pagestyle{empty}

%%%%%%%%%%%%%%%%%%%%%%%%%%%%%%%%%%%%%%%%%%%%%%%%%%%%%%%%%%%%%%%%%%%%%%%%%%%%%%%%
\begin{abstract}
As labor shortages and productivity stagnation increasingly challenge the construction industry, automation has become essential for sustainable infrastructure development. This paper presents an autonomous payload transportation system as an initial step toward fully unmanned construction sites. Our system, based on the CD110R-$3$ crawler carrier, integrates autonomous navigation, fleet management, and GNSS-based localization to facilitate material transport in construction site environments. While the current system does not yet incorporate dynamic environment adaptation algorithms, we have begun fundamental investigations into external-sensor based perception and mapping system. Preliminary results highlight the potential challenges, including navigation in evolving terrain, environmental perception under construction-specific conditions, and sensor placement optimization for improving autonomy and efficiency. Looking forward, we envision a construction ecosystem where collaborative autonomous agents dynamically adapt to site conditions, optimizing workflow and reducing human intervention. This paper provides foundational insights into the future of robotics-driven construction automation and identifies critical areas for further technological development.

\end{abstract}

%%%%%%%%%%%%%%%%%%%%%%%%%%%%%%%%%%%%%%%%%%%%%%%%%%%%%%%%%%%%%%%%%%%%%%%%%%%%%%%%
\section{INTRODUCTION}

%In the aftermath of World War II, Japan experienced a period of unprecedented economic growth and development. During this high-growth era that spanned the 1960s through the 1980s,

In the era spanning the $1960$s through the $1980$s, Japan experienced a period of unprecedented economic growth and development where it undertook massive infrastructure projects, laying the foundation for modern Japan's transportation networks, utilities, and urban landscape. 
%However, the coming years present a significant challenge. By 2033, 
However, projections indicate that by $2033$, approximately $63\%$ of Japan's bridges and $42\%$ of its tunnels will exceed their intended 50-year lifespan \cite{MLIT2022}. Its vulnerability to natural disasters accelerates structural deterioration and increases maintenance demands, further exacerbating this problem.

Compounding this challenge is Japan’s well-documented demographic crisis, as well as its astonishingly low percentage of immigrants. 
%The nation faces a rapidly aging population and a shrinking workforce, with the construction industry particularly hard-hit. 
Between $1997$ and $2023$, the construction workforce fell dramatically from $6.85$ million to just $4.83$ million workers \cite{JapanTimes2024}. In a world where unemployment continues to increase, the active job openings to applicant ratio for the construction industry was an alarming $5.86$ and is projected to increase further \cite{MLITWhitePaper}.
%This imbalance stems from several factors, most notably the industry's notorious "3K" reputation: \textit{kitsui}, \textit{kitanai} and \textit{kiken}, which roughly translate to \textit{challenging or tough}, \textit{dirty}, and \textit{dangerous}.

The severity of this situation is further highlighted due to the labor productivity crisis in the construction industry, which, unlike manufacturing and other sectors, has remained relatively stagnant over the years. According to reports, the labor productivity index for the industry has been decreasing consecutively for the last three years, most recently by $1.8\%$ \cite{JPC2025}. This productivity gap, combined with labor shortages, creates critical bottlenecks in addressing infrastructure maintenance backlogs, as well as planning future development.

In response to these challenges, the government introduced the i-Construction initiative as a strategic effort to revolutionize the construction industry through information and communication technology (ICT). Following its success, i-Construction $2.0$ was launched in $2024$ to further increase productivity by $50\%$ and reduce the number of workers on construction sites by $30\%$ by fiscal year $2040$ \cite{iConstruction2}.

In this paper, drawing from field experiences, challenges faced, and lessons learned, we present our findings and highlight research topics that form bottlenecks to the large-scale automation of the construction industry.
%Uncomment if there is enough space
%Section \ref{tech} discusses existing technologies that automate certain processes of the construction pipeline. In Section \ref{fieldwork}, we discuss the current status of our system and compile the problems faced and our learnings in Section \ref{lessons}. We propose the future prospects of a fully automated construction site in Section \ref{future} and conclude our paper with Section \ref{conclusion}.

\begin{figure}
\centering
\includegraphics[width=3in]{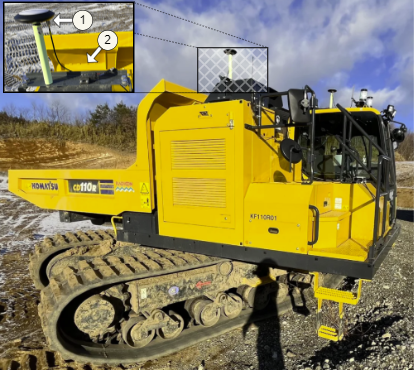}
\caption{CD110R-$3$ Crawler Carrier and Hardware components on it: a GNSS antenna $(1)$, an IMU $(2)$.}%, Remote Stop Receiver Antenna (4), Status Indicator Lamps (5).}
\label{fig_on_cab}
\end{figure}

\section{EXISTING TECHNOLOGIES} \label{tech}

The lifecycle of a civil infrastructure construction project, such as a highway or a river embankment, follows a structured but dynamic sequence of phases that must adapt to both planned requirements and unforeseen site conditions. The project begins with a comprehensive site investigation where the soil conditions, hydrological features, and surrounding infrastructure are evaluated. For highway construction, this includes assessing existing traffic patterns and environmental impact zones, whereas embankment projects require a detailed analysis of the flood plains and soil stability. This preliminary data dictates the subsequent design and engineering phase, where detailed specifications and construction drawings are developed in accordance with safety and environmental regulations.

Next is the site preparation phase, where the vegetation is cleared, access roads are established, erosion control measures are implemented and some temporary structures are set up. Following site preparation, the earthwork phase begins. Heavy machinery, including excavators, bulldozers, dump trucks, etc., move significant volumes of soil - cutting the high and filling the low areas to establish the designed grade profiles. Throughout this process, rigorous compaction testing is required to meet density specifications, critical for structural integrity.

Although subsequent phases differ significantly based on the project type, almost all projects culminate in finishing work, including drainage systems, guardrails, signage, environmental mitigation measures, and site restoration. Throughout all phases, continuous quality control testing, environmental monitoring, and safety inspections occur in parallel with the construction activities. The entire process operates under strict scheduling constraints, that must account for seasonal windows, material delivery timelines, equipment availability, and workforce allocation - creating a complex orchestration of resources that must adapt to changing site conditions, weather events, and unforeseen subsurface challenges. Recent advances in sensor technology, vehicle autonomy and artificial intelligence have helped successfully automate some phases of this medley of processes.

%\subsection{Remote Operations}

Teleoperation systems represent one of the most widely adopted technological approaches, significantly improving worker safety by removing operators from hazardous environments. However, these systems fail to address productivity challenges, as teleoperators 
%- being human,
often demonstrate lower efficiency than on-site operators working directly with the equipment \cite{LEE2022104119}. Moreover, technical limitations such as network latency affect accuracy and response time. The need to transmit high-bandwidth data streams such as images, point clouds, etc., introduces delays that impair operator decision-making capabilities. This increases the risk of damaging existing infrastructure. Perhaps most critically, these solutions become completely ineffective in environments lacking reliable network connectivity.

%\subsection{Surveillance}
% Is this part needed?
%Surveillance technologies have been integrated into construction sites to improve the safety of workers, with solutions ranging from autonomous guided vehicles (AGVs) being deployed for site monitoring \cite{braga2024}, to utilizing YOLO v8-based computer vision models on larger sites \cite{yolov8safety}.

%\subsection{Data Collection \& Mapping}

The usage of drone-based systems for mapping the job site frequently and to monitor and analyze the work progress has gained a lot of traction for improving construction planning and site analysis. Using standard techniques such as Structure from Motion (SfM), the drone images are being combined with $4$D BIM to serve as a great tool for analyzing the progress of the project \cite{bimsfm}. More recently, advances in algorithms like Neural Radiance Fields (NeRFs) and 3D Gaussian Splatting offer a cheap and scalable way to generate the model of a large construction site for any contractor to keep track of \cite{dragon}, \cite{3DGS}.

%\subsection{Automated Machinery}

Some have taken it a step further and have automated operation-specific machinery. An autonomous excavator system equipped with multiple sensors and a hierarchical planning and control system has been deployed in the real world on an actual customer site \cite{aes}. This system has further been upgraded into an end-to-end system by integrating the multi-modal sensor data with imitation learning to execute various excavation tasks \cite{exact}. The i-Construction $2.0$ initiative has also opened up new avenues in Japan, with many major companies coming into the fray. Kajima's A4CSEL \cite{kajima2024} and Taisei's T-iCraft \cite{taisei2022} are just to name a few.

Despite these advances, the construction industry lacks generalized automation systems comparable to frameworks like Tier IV's Autoware \cite{kato2018autoware} and Baidu's Apollo \cite{apollo} in the autonomous vehicles domain - solutions that can be universally applied across vehicle types. Also, the existing vehicle-specific solutions have not been seamlessly integrated into other construction processes such as surveillance, mapping and construction planning.

\begin{figure}
\centering
\includegraphics[width=3in]{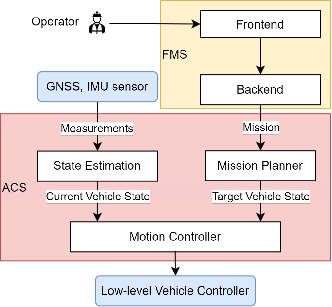}
\caption{High-level System Architecture}
\label{fig_fms_acs}
\end{figure}

\section{FIELD WORK: CURRENT STATUS} \label{fieldwork}
\subsection{System Overview}
As part of the solution to automate the entire construction pipeline, we are initially developing an autonomous payload transportation system starting with the CD110R-$3$ crawler carrier as shown in Fig. \ref{fig_on_cab} manufactured by Komatsu. The vehicle is capable of achieving a maximum speed of $10$ km/h and features a $360$-degree rotatable upper body, enabling lateral dumping and facilitating high-efficiency operations without requiring additional space for turning. Additional sensors and hardware to facilitate development of autonomous features are installed as described in Fig. \ref{fig_on_cab}.
%, Fig. \ref{fig_in_cab}.

The proposed system incorporates a custom user interface (UI) for workflow configuration and real-time system monitoring. 
In the initial development phase, external sensor-based safety features were deliberately omitted to balance the ease of site control, as compared to on-road vehicles, with expedited deployment timelines. 
Instead, a safety mechanism has been implemented wherein the fleet management system (FMS) issues a pause command to the vehicle upon detecting dynamic object intrusions into predefined operational zones. This detection is achieved by monitoring positional data from individual GNSS devices assigned to machines and personnel. Furthermore, most personnel on site are equipped with reliable remote stop buttons as an additional safety measure.

%\begin{figure}
%\centering
%\includegraphics[width=3in]{figure/in_cab.png}
%\caption{Hardware Components inside the Cabin: Additional PCB for sensors and controllers (1), Remote Stop Receiver (2), Router (3), Logging controller (4) ACS controller (5).}
%\label{fig_in_cab}
%\end{figure}

% \begin{figure}
% \centering
% \includegraphics[width=3in]{figure/fms_ui.png}
% \caption{UI of FMS: Left side of the image shows the screen of mobile application. Right side shows the web application.}
% \label{fig_fms}
% \end{figure}

%\subsection{Rationale for Prioritizing 'Transporting' Tasks}
\subsection{Why 'Transporting'?}
The short-term objective of this project is to deploy autonomous heavy machinery—albeit semi-automated—that demonstrably enhances construction site efficiency. This deployment also aims to incrementally increase automation levels while collecting multi-modal data critical for identifying workflow bottlenecks and areas for improvement. Consequently, the research and development efforts have been initially concentrated on the 'transporting' task within the broader workflow. This focus was selected due to the applicability of existing autonomous driving technologies, which accelerates development timelines. Additionally, transportation tasks inherently allow machines to encounter diverse events and objects within the operational environment, providing valuable data for further optimization.

\subsection{System Architecture}
Our system comprises multiple components, including human operators, an autonomous control system (ACS), and a fleet management system (FMS) as shown in Fig. \ref{fig_fms_acs}. The FMS has two primary modules:
%a front-end interface designed for user interaction and a back-end responsible for processing user inputs and coordinating ACS operations.

\subsubsection{FMS Front-End}
The UI enables human operators to define transportation workflows by specifying controlled vehicles, permissible operational zones. During active construction operations, the UI also facilitates real-time status monitoring, pausing and resuming of operations, and initiating vehicle transitions on designated routes. To ensure accessibility and usability across diverse roles on job sites, the UI is available as both a web-based application and a mobile application. 
%Fig.\ref{fig_fms} shows the UI of the FMS during an operation where the autonomous dump carrier is traveling to the green loading area from the blue parking area.

\subsubsection{FMS Back-End}
This module primarily handles global path planning based on user-defined parameters using the Reeds-Shepp algorithm in two-dimensional space. It generates action lists that are transmitted as missions to the ACS. The back-end also accounts for potential interferences among moving objects—including autonomous vehicles—by approximating their dimensions with safety-margin-enclosed circles. For safety assurance, it provides real-time interference detection capabilities and issues pause commands when necessary. Additionally, it monitors heartbeats from all connected agents in the site to ensure proper status updates; failure to receive these signals triggers a remote stop mechanism.

\subsubsection{ACS Mission Planner}
Tasks received from the FMS are executed here by interfacing with processes that directly control vehicle operations. The system also implements remote stop protocols in response to sensor or vehicle malfunctions. Issues like temporary connection instability lead to recoverable stops, whereas more serious issues, such as hardware failure, or localization errors, result in non-recoverable stops requiring system restart. The mission planner continuously monitors potential failures and executes appropriate responses accordingly. While current ACS operations adhere strictly to FMS-defined movement sequences, future developments aim to introduce higher levels of autonomy in the decision-making. 
%Uncomment if there's enough space (highest priority)
%For instance, combining upper-body rotation with forward travel may optimize movement efficiency compared to following longer paths without rotation.
%For instance, integrating the upper-body rotation of the CD110R-3 while planning would result in more optimal paths.

\subsubsection{ACS State Estimation}
Vehicle localization is achieved through classical sensor fusion algorithms, integrating MEMS IMU data with RTK GNSS measurements. A unique challenge arises from the inability to directly measure the relative angle between the vehicle's upper and lower body and, in some cases, the payload weight as well. 
%It is primarily estimated using IMU data, while drift corrections are applied based on Boolean signals from the vehicle, indicating alignment at $0$ or $180\degree$.

\subsubsection{ACS Motion Controller}
The motion controller consists of a subsystem for path-following control and another for upper-body rotation control.
Longitudinal path-following maintains constant speed until nearing the final point of a path segment; at this stage, speed is gradually reduced to ensure that
%Euclidean distance errors between target and actual positions remain below 50 cm.
the goal error remains below $50$ cm. 
Lateral control employs the Pure Pursuit algorithm due to its simplicity and suitability for low-speed vehicles \cite{self_driving_control_survey}. Additionally, the constant curvature of a segment in the path generated by Reeds-Shepp algorithm is known to be beneficial for stability, as it leads to a constant target angular velocity. The upper-body rotation is governed by a PID controller, with the target angles provided as mission parameters by the FMS.

\subsection{Simulation}
Given the safety concerns of testing heavy machinery, the system was thoroughly tested in a simulation environment built in Gazebo using ROS$2$ before being tested in the field. We used a replica of the actual test site, converting a point cloud created from drone images into a mesh that could be imported into Gazebo.
Multiple test cases were devised and a continuous integration (CI) workflow was set up to cover all possible scenarios.

\subsection{Field Experiments for Perception}

While thorough integration tests for evaluating the FMS-ACS interface are underway, we also conducted a basic investigation of various algorithms using the external sensors attached to the system for further advancement. The results of this investigation are reported below:

\subsubsection{Sensors \& Hardware}
Our sensor suite consists of a Tier IV C1 camera, a LiDAR, and an IMU. We investigated three LiDARs with different point cloud densities—Livox Mid$360$, Hesai XT-$32$ and a Livox HAP—to select the one that best fits our requirements. Although the Livox HAP surpasses the others in terms of point cloud density, the $360\degree$ FoV of the Hesai XT-$32$ was more appealing due to its additional usability in localization. The sensor setup was mounted at different heights in simulation to assess the FoVs, blind spots and occlusions caused by moving parts. Effects of vibration and maintenance requirements due to mud splashes were also considered before determining a temporary mounting setup.
%as shown in Fig.\ref{sensors}.

%Advantech AFE-R770 Controller for logging (4), Advantech TREK60 for ACS controller 

%Considering the ruggedness required, we use the Advantech TREK60-FL as the main controller. 
%Rated for operations in temperatures ranging from $-40\degree$ to $85\degree$, and having been tailored for heavy-duty industrial and off-road applications according to EN$60721-3-5$ class $5$M$3$ standards, we found the Advantech TREK$60$ to be most suitable for usage in the harsh construction environments. To distribute the load, we also used a separate controller specifically for data logging purposes.

\begin{figure*}[!ht]
\centering
\includegraphics[width=6.3in]{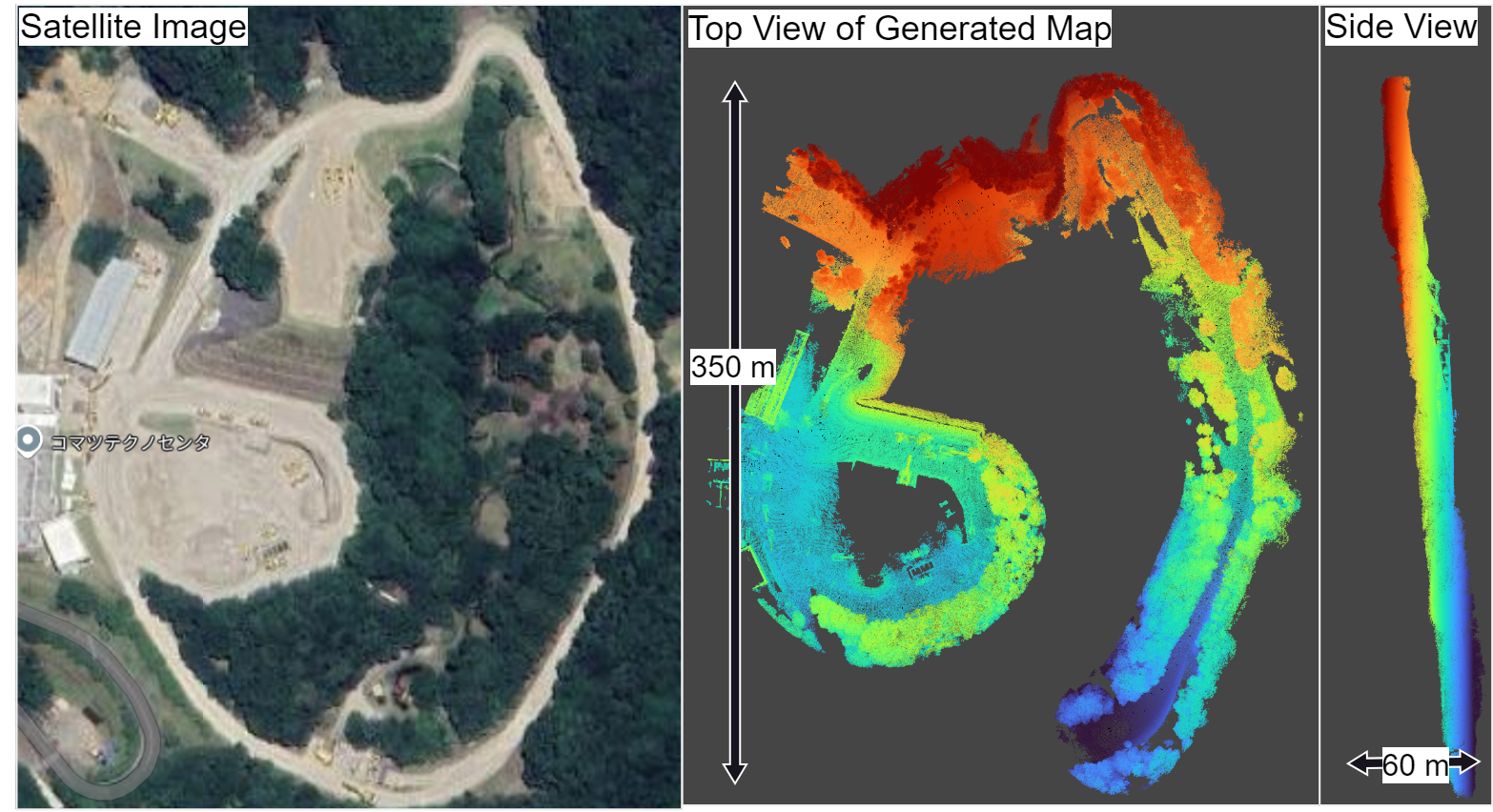}
\caption{Generated 3D Point Cloud Map (colored by elevation)}
\label{slam_ito}
\end{figure*}

\begin{figure*}[t]
\centering
\includegraphics[width=6.3in]{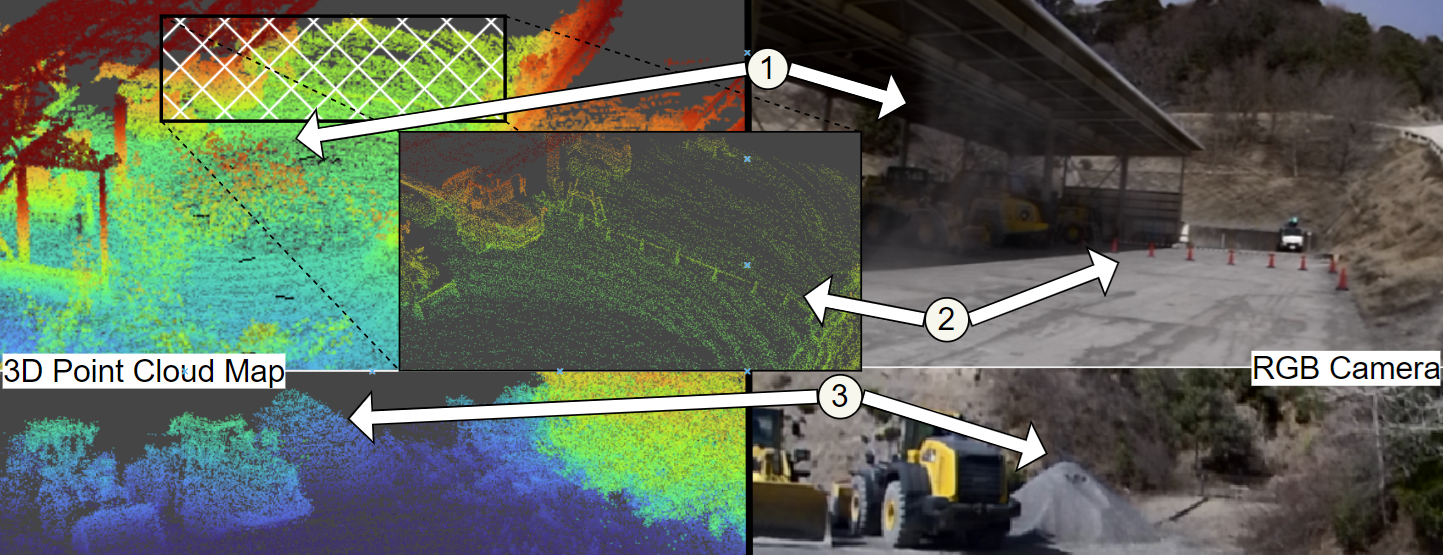}
\caption{Closer look of the map: The whirlwind with sand dust (1), and the safety cones (2) get mapped. The pile of dirt (3) also gets mapped. Either temporary features should not be mapped or the map needs to be updated in short intervals.}
\label{slam_objects}
\end{figure*}

\subsubsection{SLAM}
To enhance our system's ability to map the surrounding environment precisely and localize itself even in GNSS-denied situations, we processed the collected data using GLIM \cite{KOIDE2024104750}, an open-source SLAM software. Fig.\ref{slam_ito} shows a merged map of four logging sessions, using the Hesai XT-$32$ LiDAR and an external IMU, based on global optimization with the constraint of overlapping sub-maps. Although a quantitative evaluation of the results is still ongoing, the mapping performance appears promising.

However, as illustrated in Fig.\ref{slam_objects}, we noticed some interesting objects—other than heavy machinery—that are not being detected but need to be removed from the map. 
%The first is the whirlwind with sand dust. Even though an outlier algorithm is already applied in the pre-processing step of SLAM, similar points are observed in various locations of the map. 
First, fugitive dust clouds appear at various locations in the map despite the implementation of an outlier algorithm in the pre-processing step of SLAM.
This might be problematic in windier conditions depending on weather or location. 
%The second is the cones in Fig.\ref{slam_objects} and also various tentative objects like a stand of surveying equipment, drone and so on in other locations. 
Secondly, temporary installations such as cones, survey equipment, and drones also appear in some places.
%These would not heavily affect localization accuracy using prior map, but if such things are mapped on a way of vehicles, they may plan a route to avoid it, which may no longer exist. 
While these transient elements do not substantially impact localization accuracy when using prior maps, they can introduce inefficiencies when included in global and local maps for navigation.
%The last one is the pile of dirt, which is likely to have changed its shape soon, by machines. 
Lastly, dirt piles created by vehicle operations are likely to be reshaped by other active vehicles nearby.

%It is better to not map, or update them in short period of time as they are not permanent features of the environment

\begin{figure*}[ht!]
\centering
\includegraphics[width=6in]{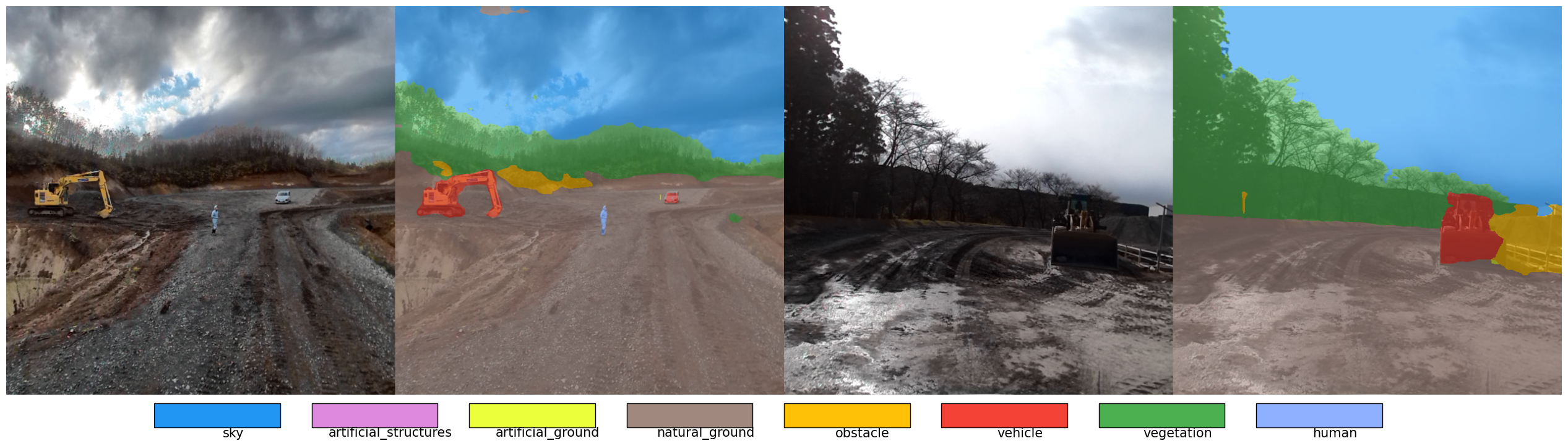}
\caption{Samples from our test data and corresponding semantic segmentation. All test data are collected from unstructured environment with the camera, the semantic labels are the inference results of our segmentation model.}
\label{image_seg}
\end{figure*}

\begin{figure*}[ht!]
\centering
\includegraphics[width=6in]{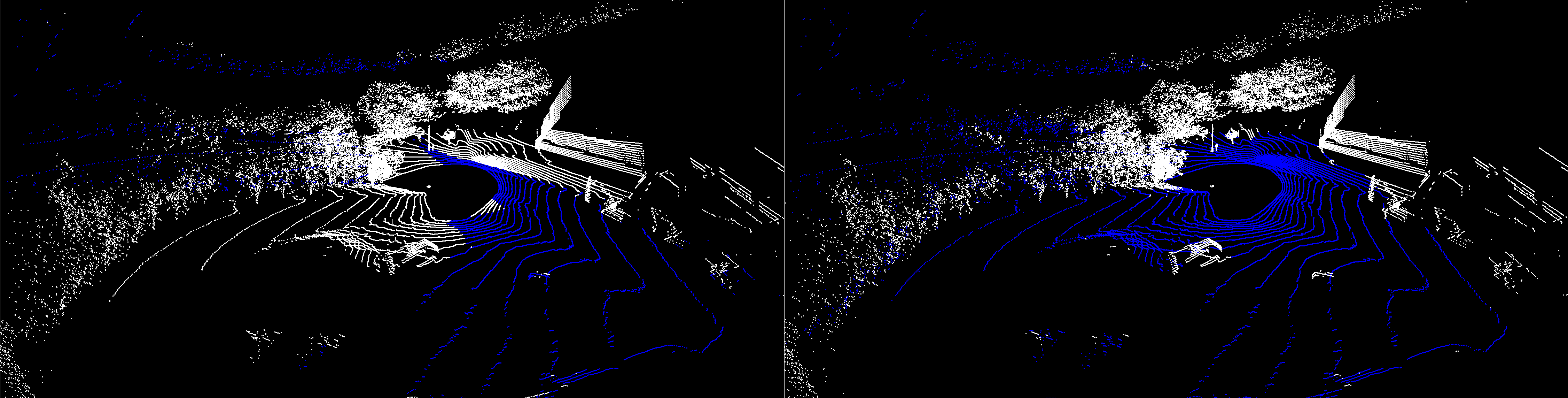}
\caption{Samples of ground segmentation of LiDAR data collected, with PTv3 (right) and fast segmentation (left).}
\label{lidar_seg}
\end{figure*}

\subsubsection{Perception}
We are also developing a perception system with a primary focus on object detection and ground segmentation. Leveraging multiple open-source off-road driving datasets, such as GOOSE \cite{goose}, \cite{gooseex}, and RELLIS-3D \cite{rellis3d}, we try to establish a baseline for perception systems operating in construction site environments. 
%Beyond utilizing existing datasets, we have actively collected proprietary perception data from multiple test sites to enhance domain-specific training.
Due to observed dissimilarities, we also actively collected data from multiple test sites to enhance domain-specific training.

The current system utilizes camera data for 
%obstacle detection, dynamic object identification, and tracking, 
object detection, while LiDAR sensor data is used to extract ground surface terrain information.
%We perform 2D semantic segmentation and object detection on sequential camera frames. 
%Multiple machine learning model architectures have been explored and compared with respect to the accuracy/performance tradeoff inherent in resource-constrained systems. As experimental evaluations are still ongoing, definitive conclusions regarding optimal model selection remain pending. 
Qualitative samples of our segmentation results using PIDNet \cite{xu2022pidnet} are presented in Fig. \ref{image_seg}, demonstrating the system's capabilities in construction site environments. However, due to the extensive variety of objects within the obstacle class, our model exhibits susceptibility to false positive detections, as shown.

For ground segmentation of LiDAR data, we evaluated several existing algorithms and pre-trained large-scale models. Fig. \ref{lidar_seg} presents comparative results from two approaches: Fast Segmentation \cite{7989591} and PointTransformer V3 (PTv3) \cite{wu2024ptv3}.
%The first algorithm implements a methodology based on research in fast segmentation of 3D point clouds \cite{7989591},  which extracts ground surfaces through an iterative process using deterministically assigned seed points. 
%This approach achieves efficient runtime performance with low computational complexity. 
%The second approach we implemented is the PointTransformer V3 (PTv3) model \cite{wu2024ptv3} pre-trained on GOOSE and GOOSE-Ex \cite{gooseex} datasets. 
%Given that this approach has effectively learned and generalized from unstructured terrain patterns, it demonstrates superior performance on construction site ground segmentation tasks. 
Trained on the GOOSE and GOOSE-Ex \cite{gooseex} datasets, PTv3 demonstrates superior performance in construction site ground segmentation tasks.
However, while Fast Segmentation achieves efficient runtime performance with low computational complexity, PTv3's runtime performance on edge devices fails to meet real-time inference requirements despite its notable accuracy advantages.
To address these limitations, further experiments will be conducted to identify approaches better suited for uneven terrain classification.
%, while simultaneously exploring the development of lightweight machine learning models incorporating knowledge distillation techniques \cite{hinton2015distillingknowledgeneuralnetwork}. 

\section{LESSONS LEARNED} \label{lessons}

\subsection{Sensors \& Hardware}

\begin{figure*}[!]
    \centering
    % All images in one row
    \subfloat[Too few distinct features]{%
        \includegraphics[width=0.245\textwidth]{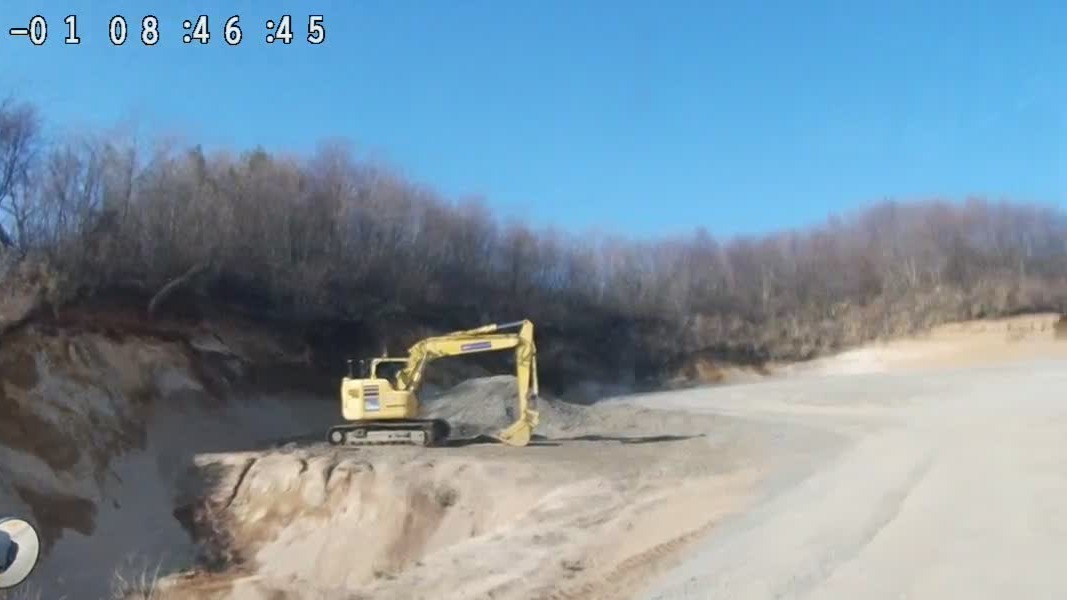}%
        \label{fig:fig1}%
    }%
    \hfill
    \subfloat[Traversable area difficult to detect]{%
        \includegraphics[width=0.245\textwidth]{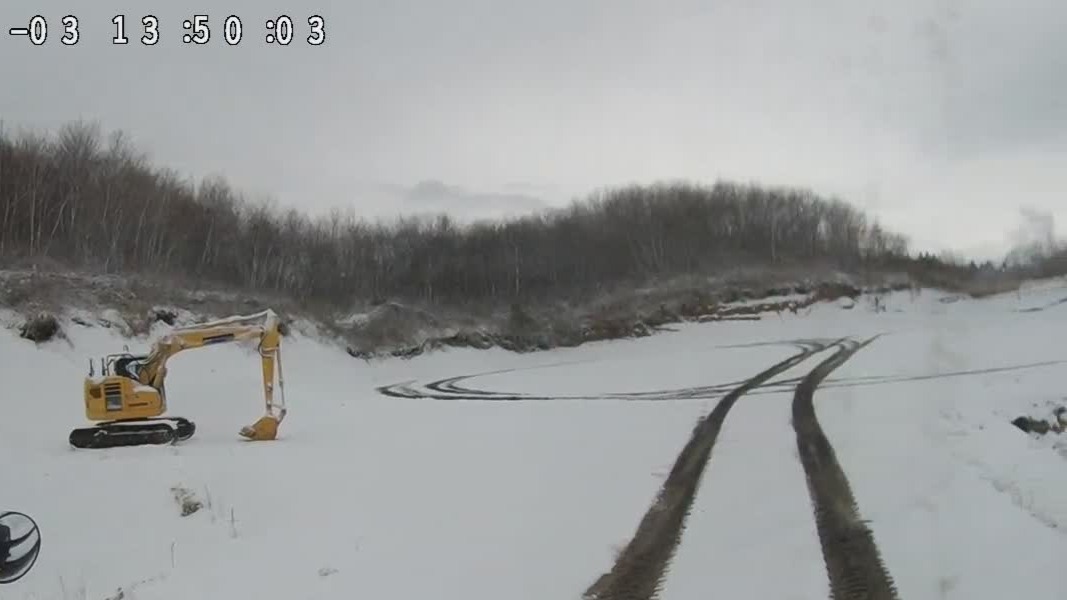}%
        \label{fig:fig2}%
    }%
    \hfill
    \subfloat[Sensors can get too dirty]{%
        \includegraphics[width=0.245\textwidth]{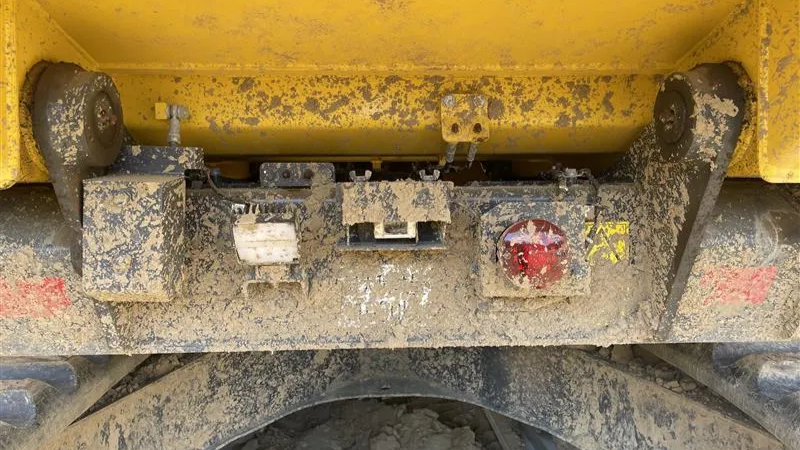}%
        \label{fig:fig3}%
    }%
    \hfill 
    \subfloat[Risky maneuvers at high speeds]{%
        \includegraphics[width=0.245\textwidth]{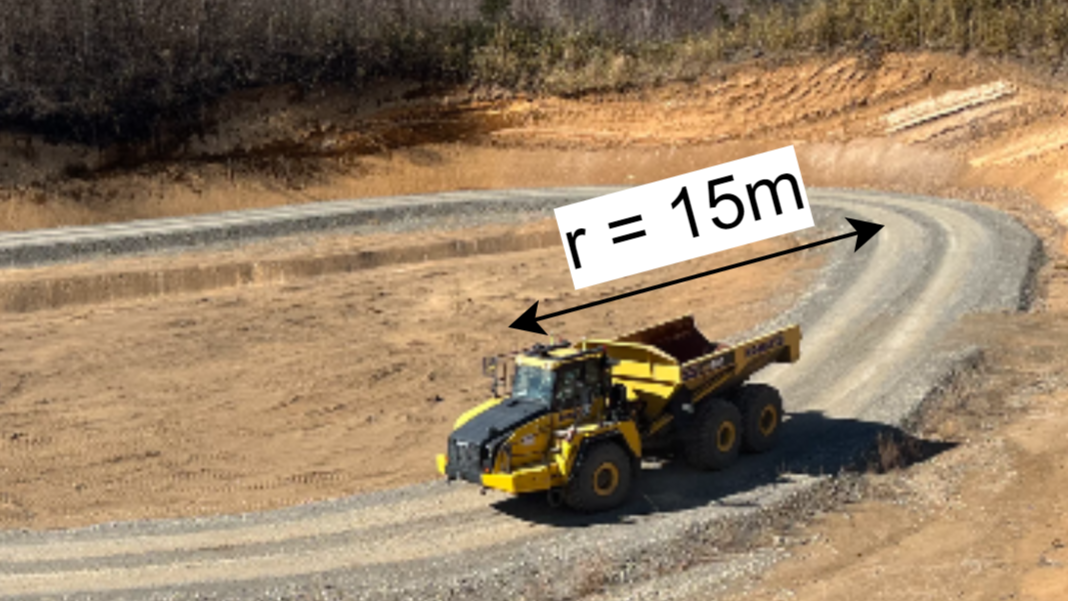}%
        \label{fig:fig4}%
    }%
    \caption{Some unique perception challenges on construction sites}
    \label{fig:all}
\end{figure*}

While a Tesla, a vehicle measuring $4.7$x$1.85$x$1.44$ meters, is equipped with eight cameras to safely navigate on roads, construction machinery like the CD110R-$3$, spanning $6.0$x$2.88$x$3.2$ meters, demands an exponentially more adept sensor infrastructure. Sensors placed too high leave blind spots around the machine, raising safety concerns, while those placed too low become susceptible to mud splashes as shown in Fig.\ref{fig:fig3}, necessitating frequent cleaning and maintenance. When positioned at an intermediate height, moving parts—such as an excavator's arm—can obstruct portions of the view, making it difficult to get a $360\degree$ view of the surroundings. The differences in vibrational characteristics, further limit optimal sensor placement.

The sheer scale and dynamic nature of these vehicles necessitates a comprehensive sensing and hardware solution that can operate under the extremest of environmental and operational conditions. However, incorporating multiple sensor types introduces another challenge: calibration. Machines with multiple moving parts navigating rough terrain require advanced self-calibration techniques that can dynamically adjust extrinsic parameters to maintain safety-critical accuracy.

Higher number of sensors and advanced fusion techniques increase the computational load on edge controllers, creating hurdles in temperature regulation, power management, as well as resource allocation. These factors compel engineers to develop innovative mounting solutions that can withstand extreme mechanical and environmental stress while simultaneously maintaining sensor integrity.

\subsection{Localization \& Mapping}

Arguably the most important module in an autonomous system, localization forms the basis for both navigation as well as perception system. While GNSS is generally available in open ground, it is often unreliable. Considering the safety of the systems as well as of the job site, it is important to fuse it with other sensors like LiDARs and IMUs. However, this brings its fair share of troubles. 

Unlike controlled urban environments with fairly constant reference points, construction sites are defined by their dynamic nature.
%- rapidly changing terrains, temporary structures and continuously transforming operational zones. 
For large-scale infrastructure projects that extend for months to even a couple of years, the changes caused by passing seasons can completely confound a camera or LiDAR-based localization module. The long duration operations also introduce sensor drift, leading to inaccuracies in positioning that accumulate over time, potentially causing navigation errors.

Especially for construction sites, the need for precise mapping goes beyond merely navigation purposes. Along with the static maps, the incremental changes in the site as construction progresses can be used as a critical tool for understanding and optimizing material movement, resource requirement and project development.

\subsection{Navigation}

Controlling autonomous heavy machinery on construction sites, often characterized by unforeseeable environmental variability and frequent terrain changes, represents a complex engineering challenge. Unlike environments such as warehouses or manufacturing floors, construction sites are inherently dynamic landscapes that transform continuously. Temporary structures, moving equipment, and the changing terrain mean that global mapping approaches become quickly obsolete. Let alone other factors, the mass of the vehicle itself is capable of altering the ground enough to warrant a new map. Either the global map needs to be updated, sometimes even hourly, or the local mapping systems need to be deft enough to adapt to the rapid environmental changes.

Construction sites often feature diverse soil conditions - ranging from soft clay to rocky substrates, each presenting a unique challenge for vehicle planning and control. Weather conditions further complicate navigation, with precipitation, temperature fluctuations, and seasonal changes dramatically altering ground conditions, soil stability and vehicle traction parameters. Considering this, along with steep gradients and challenging topological features, precise control becomes important to prevent dangerous situations.

% Can be removed if needed
%Operations of certain machinery also require near-surgical precision: bulldozers must flatten the terrain to tight tolerances, while excavators must control the arm with millimeter-level accuracy when operating near critical infrastructure like drainage systems.

Due to the extreme nature of the surroundings and the high level of precision involved, proper tuning of hyperparameters also becomes important. However, these values have been found to depend on an interplay of various factors like payload mass, soil and terrain characteristics, weather, and potentially even temperature-induced variations in hydraulic fluid viscosity. Perhaps the most critical challenge lies in balancing operational efficiency with safety. Heavy construction machinery like the HM$400$ articulated dump truck—weighing upto a maximum of $75$ metric tonnes—can achieve maximum speeds of $30$-$50$ km/h and travel through curves with minimum turning radius of $8.5$m. Autonomously navigating such massive machines through tight, obstacle-laden spaces at these velocities introduces a significant risk potential. Thus, autonomous systems for construction machinery must develop sophisticated algorithms that can dynamically adjust speed, trajectory, and operational parameters in real-time, prioritizing worker safety without compromising efficiency.

\subsection{Environmental Perception}

Understanding the surroundings, the obstacles in it, and the vehicle's position relative to them becomes very important for safe and reliable motion. Like the previous modules, this too comes with its own set of challenges.

Fugitive dust clouds, a ubiquitous feature of construction environments, create significant interference with both LiDAR point cloud data and camera systems. These particulate clouds not only obscure sensor information but also generate complex scattering effects that can distort both navigation and perception systems.

Visibility constraints compound these challenges, with lighting conditions ranging from harsh direct sunlight to dark, shadowy areas that can severely affect sensor performance. Weather conditions like fog can make RGBD cameras obsolete whereas snow or rain can seriously challenge localization and detection using LiDARs. Due to the variability of construction project locations, certain elements, like water surfaces and puddles, that would be classified as "non-traversable" or "obstacles" on some sites become "traversable" on some. Not just that, a surface that is non-traversable during one moment might become passable at a later point, such as the dirt pile in Fig. \ref{slam_objects}, demanding an unprecedented level of contextual understanding from perception systems.

%\subsection{FMS \& Communication}

\subsection{Datasets \& Simulation}

Advances in the usage of AI for autonomous driving can be mainly attributed to the numerous datasets available and the significant improvement in the quality of simulators. However, there is a very obvious dearth of publicly available data from active construction sites.

Although off-road driving datasets (\cite{goose}, \cite{gooseex}, \cite{rellis3d}) do a good job in capturing unstructured environments, the dynamic nature of a construction site as well as the challenging conditions are not well reflected. A critical limitation of existing datasets is their crude classification of construction vehicles under the general label of \emph{heavy machinery}. This lack of granularity severely hampers the development of systems capable of understanding the complex interactions between different types of machinery on a construction site. Furthermore, current datasets struggle with environmental factors unique to construction sites, most notably fugitive dust clouds generated due to vehicle motion, material deformation, and the ever-changing terrains.

The inability to replicate and test edge cases in physical environments presents another substantial challenge. Safety considerations make it extremely difficult to test rare but critical scenarios on actual construction sites, as it would expose personnel to unacceptable risks. Simulating them with accurate vehicle dynamics in high-fidelity, photorealistic environments is thus absolutely essential to advancing construction robotics safely.

\section{FUTURE PROSPECTS} \label{future}
Even though our short-term goal, as mentioned in Section \ref{fieldwork}, is to deploy practical autonomous systems to actual job sites, we eventually aim for a completely unmanned solution that not just optimizes construction efficiency but is also adaptable to diverse projects. In order to achieve this, the ACS needs to be integrated with other technologies to realize the following key points:

\subsection{Collaborative Environmental Understanding}

One of the biggest challenges discussed in Section \ref{lessons} was the rapidly evolving terrain on the job site. This made environmental perception difficult, thus requiring a whole pipeline of collecting point cloud data, processing it, converting it into maps and then using it for motion planning. To simplify this, multiple autonomous systems can be integrated to create a continuously-updated, high-fidelity representation of the construction site. Specialized mapping agents—including drones and quadrupeds—can systematically survey designated zones of the site at regular intervals. The agility of these platforms allows them to navigate areas inaccessible to larger vehicles, capturing detailed topographical data, identifying the temporary structures installed intermittently, and simultaneously monitoring the progress across their assigned sectors. This data can be wirelessly transmitted to a centralized dashboard system, which seamlessly stitches these mapping fragments into a cohesive, site-wide model. The FMS can access this unified map to dynamically optimize vehicle routing decisions, considering not just the static layout, but also recently identified obstacles and newly completed work zones. Individual machines can also leverage the specific regions of the maps for their immediate operational needs - performing real-time detection for dynamic obstacle avoidance. This hierarchical approach can create a robust ecosystem with a consistent understanding of the rapidly changing construction environment, leading to increased efficiency in project planning.

\subsection{Maximize Construction site Efficiency}

A promising avenue for future research lies in extending automation beyond machinery operation to encompass higher-level construction management functions. Currently, construction site efficiency is heavily dependent on human decision-making for critical tasks including resource allocation, material movement planning, equipment scheduling, and construction sequencing. These planning processes, while executed by experienced professionals, remain vulnerable to cognitive limitations when optimizing complex systems with numerous, inter-dependencies and constraints.

The application of reinforcement learning and advanced AI techniques to these planning domains can significantly increase efficiency. AI-based systems have the potential to simultaneously optimize competing objectives - such as minimizing material movement distances, reducing equipment idle time, and optimizing fuel consumption - while adapting to real-time changes in site conditions. Using the comprehensive environmental data collected through collaborative mapping frameworks, these systems could generate detailed execution plans that consider not just the current state of the site, but also predictive models of how operations will unfold over time. The resulting output will be highly detailed, optimized plans for all aspects of site operations.

Whereas in Autonomy Level $3$ \cite{autonomylevels}, a human-in-the-loop paradigm can be adopted where experienced construction managers review and approve the AI-generated plans before implementation, Level $4$ and Level $5$ would move further towards complete autonomy with entire operations being handled by the system.

\subsection{Distributed Autonomous Agents}

While modern machinery is designed with automation compatibility, older vehicles often cannot be cost-effectively retrofitted. Rather than relegating these legacy systems to scrap yards, humanoids could be trained to operate these equipment, leveraging their physical form factor to interface with controls designed for human operators. This approach will enable a mixed fleet solution, thus maximizing resource utilization while facilitating a gradual transition to fully automated worksites.

Another avenue to explore is the evolution from centralized control architectures to distributed autonomy systems, where multiple heterogeneous agents operate collaboratively within designated operational zones to accomplish high-level construction tasks. Diverse machinery along with specialized robots capable of direct machine-to-machine communication and autonomous collaboration without constant reliance on a central system, will allow machines to dynamically reorganize their activities in response to changing site conditions or unexpected challenges.

Apart from construction-related activities, site monitoring, and vehicle maintenance and inspection, etc., are some critical functions which can also be performed by robots without human supervision. The integration of such support robots will address many of the maintenance challenges that currently limit long-duration operations, creating a self-sustaining eco-system of machines that can collectively maintain their operational readiness.

% \subsection{Remote Monitoring System}
% Even though if we were able to achieve near-unmanned construction workflow in actual job sites, it would still take a long time to completely eliminate human supervision.

\section{Conclusion} \label{conclusion}

In this paper, we present applications of robotics and automation in the construction and civil engineering industry. We explain the current status of our autonomous transportation system, with a special focus on the lessons learned from field experiments. We further discuss future prospects, with a vision of a completely unmanned construction site.

% \addtolength{\textheight}{-12cm}   % This command serves to balance the column lengths
                                  % on the last page of the document manually. It shortens
                                  % the textheight of the last page by a suitable amount.
                                  % This command does not take effect until the next page
                                  % so it should come on the page before the last. Make
                                  % sure that you do not shorten the textheight too much.

\section*{ACKNOWLEDGMENT}

The authors would like to thank Hirotaka Nakamura and Hiroki Nagasaki for accommodating all our requests for changes in sensor mounts and their mounting locations. The FMS wouldn't have been possible without Takeshi Yamada and Kazuma Hotta. We would further like to thank Tosuke Iwanaga for his help with the experiments and critically reviewing the paper. We also appreciate the contributors of all the open source tools used.

\bibliographystyle{IEEEtran}
\bibliography{IEEEabrv, bib}

\end{document}